\documentclass[letterpaper]{article} 

\usepackage[preprint]{aaai2027}  

\usepackage[hyphens]{url}  
\usepackage{graphicx} 
\usepackage{natbib}  
\usepackage{caption} 
\usepackage{amsmath}
\usepackage{amssymb}
\usepackage{booktabs}
\usepackage{multirow}
\usepackage{makecell}
\usepackage{listings}
\newcommand{\uline}[1]{\underline{#1}}

\newcommand{\benchname}{\textsc{WildHandBench}}
\newcommand{\logobox}[1]{\raisebox{-0.2\height}{#1}\,}
\expandafter\def\csname logoimg@baidu\endcsname{\logobox{\includegraphics[height=1.2em]{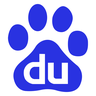}}}
\expandafter\def\csname logoimg@claude\endcsname{\logobox{\includegraphics[height=1.2em]{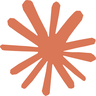}}}
\expandafter\def\csname logoimg@deepseek\endcsname{\logobox{\includegraphics[height=1.2em]{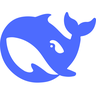}}}
\expandafter\def\csname logoimg@gemini\endcsname{\logobox{\includegraphics[height=1.2em]{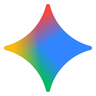}}}
\expandafter\def\csname logoimg@glmv\endcsname{\logobox{\includegraphics[height=1.2em]{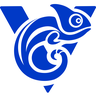}}}
\expandafter\def\csname logoimg@hunyuan\endcsname{\logobox{\includegraphics[height=1.2em]{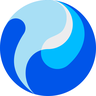}}}
\expandafter\def\csname logoimg@intern\endcsname{\logobox{\includegraphics[height=1.2em]{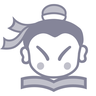}}}
\expandafter\def\csname logoimg@kimi\endcsname{\logobox{\includegraphics[height=1.2em]{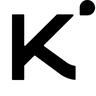}}}
\expandafter\def\csname logoimg@mineru\endcsname{\logobox{\includegraphics[height=1.2em]{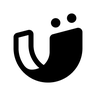}}}
\expandafter\def\csname logoimg@paddle\endcsname{\logobox{\includegraphics[height=1.2em]{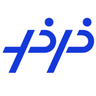}}}
\expandafter\def\csname logoimg@qwen\endcsname{\logobox{\includegraphics[height=1.2em]{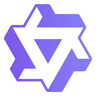}}}
\expandafter\def\csname logoimg@xiaohongshu\endcsname{\logobox{\includegraphics[height=1.2em]{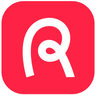}}}
\newcommand{\logo}[1]{\ifcsname logoimg@#1\endcsname\csname logoimg@#1\endcsname\fi}

\title{\benchname{}: A Benchmark for Handwritten Text Understanding that Challenges MLLMs and Humans}

\author{
    Jun Zhang,
    Qiao Zhao,
    Cheng Cui,
    Jianying Qu,
    Zhongkai Sun,
    Jianwen Yang,
    Changda Zhou,
    ZhuoXin Liu,
    Shubin Han
}
\affiliations{
    Baidu Inc.
}

\begin{document}

\maketitle

\begin{abstract}

While the top model on OmniDocBench now reaches 96.34\% overall on printed-document parsing, the ability of current models to handle challenging handwritten documents remains largely uncharacterized. Existing benchmarks focus on isolated text or formulas, overlook handwritten tables and real-world degradation, and report aggregate accuracy without explaining why models fail.

We present \textbf{\benchname{}}, a benchmark containing 500 handwritten documents across three structures (free text, tables, formulas), four languages, and nine real-world scenarios. We introduce a \textbf{Prior-Driven Error (PDE)} metric that quantifies whether errors originate from language priors rather than visual evidence. Evaluating 18 state-of-the-art models together with calibrated human baselines, we find: (1)~the best model achieves only \textbf{71.85\%} overall; (2)~humans outperform all models yet the gap is narrow (77.09\% vs.\ 71.85\%); and (3)~model errors are qualitatively different from human errors---63--91\% of model errors are prior-driven versus only 49\% for humans, exposing systematic reliance on language priors that conventional accuracy metrics cannot capture.
\end{abstract}

\section{Introduction}

Recent advances in multimodal large language models (MLLMs) have dramatically improved document understanding~\cite{ocrbenchv2,omnidocbench}. The top model on the OmniDocBench leaderboard now reaches 96.34\% overall~\cite{omnidocbench}, approaching saturation on printed-document parsing. These remarkable results have gradually created a widespread perception that OCR is becoming a largely solved problem.

We argue that this conclusion is premature.

Nearly all evidence supporting this perception comes from printed documents with relatively regular layouts and clear visual appearances~\cite{docvqa,pubtabnet,fintabnet,icdar_table}. Handwritten documents tell a different story. Medical records, handwritten forms, classroom notes, historical archives, and personal correspondence remain substantially more difficult for today's MLLMs despite their impressive performance on printed documents. Unlike printed OCR, handwritten document understanding requires models to jointly resolve ambiguous handwriting, recover irregular document structures, distinguish visual evidence from language priors, and reason under genuine uncertainty~\cite{hallusionbench,hallucination_survey}. Consequently, handwriting is not simply a more difficult OCR problem, but a distinct document understanding challenge.

Surprisingly, current evaluation benchmarks provide limited insight into why handwritten document understanding remains unsolved. Existing handwritten benchmarks primarily evaluate isolated free-text recognition, while handwritten tables, naturally written formulas, and real-world document degradation remain largely unexplored. More fundamentally, existing evaluations focus almost exclusively on recognition accuracy~\cite{ocrbenchv2, omnidocbench}. They measure \emph{how often} models fail, but provide little understanding of \emph{why} they fail---in particular, whether errors originate from poor visual perception or from excessive reliance on language priors.

To address this gap, we introduce \textbf{\benchname{}}, a benchmark for handwritten document understanding. \benchname{} contains 500 handwritten documents spanning three document structures (free text, tables, and formulas), four language settings, and nine representative real-world scenarios. We also introduce a \textbf{Prior-Driven Error (PDE)} metric that characterizes whether model errors arise from language priors instead of visual evidence, enabling systematic analysis of hallucination-like recognition behaviors.

Using \benchname{}, we conduct a comprehensive evaluation of 18 state-of-the-art MLLMs and OCR-oriented vision-language models under a unified evaluation protocol together with calibrated human baselines. Our study yields three key findings. First, handwritten document understanding remains far from solved: the strongest model achieves only 71.85\% overall. Second, humans outperform all models (77.09\% vs.\ 71.85\%), yet the gap is narrow---top models even surpass humans on text transcription under format-matching metrics. Third, model errors are qualitatively different from human errors: humans remain conservative on illegible content, whereas models confidently hallucinate fluent but unsupported text, with 63--91\% of model errors classified as prior-driven compared to only 49\% for humans. This exposes systematic reliance on language priors that conventional accuracy metrics cannot capture.

Our contributions are summarized as follows:

\begin{itemize}
\item We propose \benchname{}, to our knowledge the first benchmark that jointly evaluates handwritten free text, tables, and formulas across multiple languages and diverse real-world scenarios, addressing key gaps in existing evaluation coverage.

\item We introduce the Prior-Driven Error (PDE) metric, which quantifies whether model errors originate from language priors rather than visual evidence, enabling systematic analysis of hallucination-like recognition behaviors beyond conventional accuracy metrics.

\item We conduct a comprehensive evaluation of 18 state-of-the-art MLLMs and OCR-oriented vision-language models together with calibrated human baselines, revealing the persistent handwriting gap, the narrow yet meaningful distance to human performance, and the systematic reliance of MLLMs on language priors.
\end{itemize}

\section{Related Work}

\textbf{Handwritten document understanding.}
Handwritten document analysis has been studied for decades, yet existing benchmarks remain highly task-specific and fragmented. Most datasets focus on line-level handwritten text recognition, including IAM~\cite{iam}, RIMES~\cite{rimes}, CASIA-HWDB~\cite{casia_hwdb}, SCUT-EPT~\cite{scut_ept}, and SCUT-HCCDoc~\cite{scut_hccdoc}, covering Latin and Chinese handwriting under relatively controlled settings. Scientific handwriting has been investigated through datasets such as NoTeS-Bank~\cite{notesbank}, while handwritten mathematical expression recognition is primarily benchmarked by CROHME~\cite{crohme} and MathWriting~\cite{mathwriting}. In contrast, table understanding benchmarks, including PubTabNet~\cite{pubtabnet}, FinTabNet~\cite{fintabnet}, and the ICDAR table competitions~\cite{icdar_table}, are exclusively constructed from printed documents. Although these datasets have significantly advanced individual OCR tasks, they evaluate text, formulas, and tables independently and provide limited coverage of challenging handwritten documents with real-world degradation. Consequently, current handwritten benchmarks remain insufficient for evaluating handwritten documents as a unified document understanding problem.

\vspace{0.4em}

\textbf{Benchmarking document understanding.}
The rapid development of multimodal large language models (MLLMs) has shifted document evaluation from isolated OCR tasks toward holistic document understanding. Benchmarks such as OmniDocBench~\cite{omnidocbench} evaluate diverse document parsing capabilities under a unified framework, while Real5-OmniDocBench~\cite{real5omnidocbench} further investigates robustness under real-world physical degradations through fine-grained factor-level evaluation. OCRBench~\cite{ocrbench} and OCRBench v2~\cite{ocrbenchv2} extend benchmark coverage to OCR-oriented multimodal capabilities, including text recognition, formula parsing, table understanding, and document visual question answering. These benchmarks demonstrate remarkable progress on printed documents, with the top model now reaching 96.34\% overall on OmniDocBench~\cite{omnidocbench}. Nevertheless, these evaluations primarily report aggregate recognition accuracy, providing limited understanding of why handwritten documents remain substantially more challenging than printed documents.

\vspace{0.4em}

\textbf{Prior-driven errors in vision-language models.}
Recent studies have shown that multimodal large language models (MLLMs) frequently generate outputs that are not fully grounded in visual evidence~\cite{hallusionbench,hallucination_survey}. In OCR-related tasks, such hallucinations often manifest as fluent yet visually unsupported transcriptions, where language priors override ambiguous visual observations. To better characterize this phenomenon, Seong \emph{et al.}~\cite{vlm_overcorrect} proposed PINK, which penalizes over-correction in handwritten mathematical expression recognition. Similarly, HunyuanOCR-1.5~\cite{hunyuan_ocr15} introduces CHAOS-Bench, which evaluates whether models faithfully preserve visually observed but semantically implausible text under synthetic character perturbations in printed documents. These studies consistently demonstrate that prior-driven recognition errors are systematic rather than incidental.

However, existing evaluations remain restricted to either handwritten mathematical expressions or synthetically perturbed printed documents. To the best of our knowledge, no existing benchmark systematically measures prior-driven errors on naturally occurring handwritten documents spanning free text, tables, and formulas. Our proposed Prior-Driven Error (PDE) metric is designed to bridge this gap by quantifying the extent to which model errors are attributable to language priors instead of visual evidence in realistic handwritten document understanding.

Table~\ref{tab:comparison} summarizes representative benchmarks. Compared with previous datasets, \benchname{} is, to our knowledge, the first to unify handwritten free text, tables, and formulas within a single evaluation framework with calibrated human baselines and Prior-Driven Error analysis.

\begin{table*}[!ht]
\centering
\small
\setlength{\tabcolsep}{4pt}
\begin{tabular}{@{}lccccccc@{}}
\toprule
Benchmark & Lang. & Handwritten & Free Text & Table & Formula & Real Degrad. & Human Baseline \\
\midrule
IAM & EN & \checkmark & \checkmark & -- & -- & -- & -- \\
CASIA-HWDB & ZH & \checkmark & \checkmark & -- & -- & -- & -- \\
SCUT-EPT / HCCDoc & ZH & \checkmark & \checkmark & -- & -- & Partial & -- \\
CROHME / MathWriting & Symbol & \checkmark & -- & -- & \checkmark & -- & -- \\
PubTabNet / FinTabNet & EN & -- & -- & \checkmark & -- & -- & -- \\
ICDAR Table & EN & -- & -- & \checkmark & -- & -- & -- \\
OCRBench (v1/\,v2) & Multi & Partial & \checkmark & \checkmark & \checkmark & Partial & -- \\
OmniDocBench & ZH/EN & Partial & \checkmark & \checkmark & \checkmark & Partial & -- \\
Real5-OmniDocBench & ZH/EN & Partial & \checkmark & \checkmark & \checkmark & \checkmark & -- \\
\midrule
\benchname{} (Ours) & ZH/EN & \checkmark & \checkmark & \checkmark & \checkmark & \checkmark & \checkmark \\
\bottomrule
\end{tabular}
\caption{Comparison with representative benchmarks. \benchname{} is, to our knowledge, the first to jointly cover handwritten free text, tables, and formulas with real-scenario degradation and calibrated human baselines.}
\label{tab:comparison}
\end{table*}

\section{\benchname{}: Benchmark Design}

\begin{figure*}[!ht]

\centering

\includegraphics[width=\textwidth]{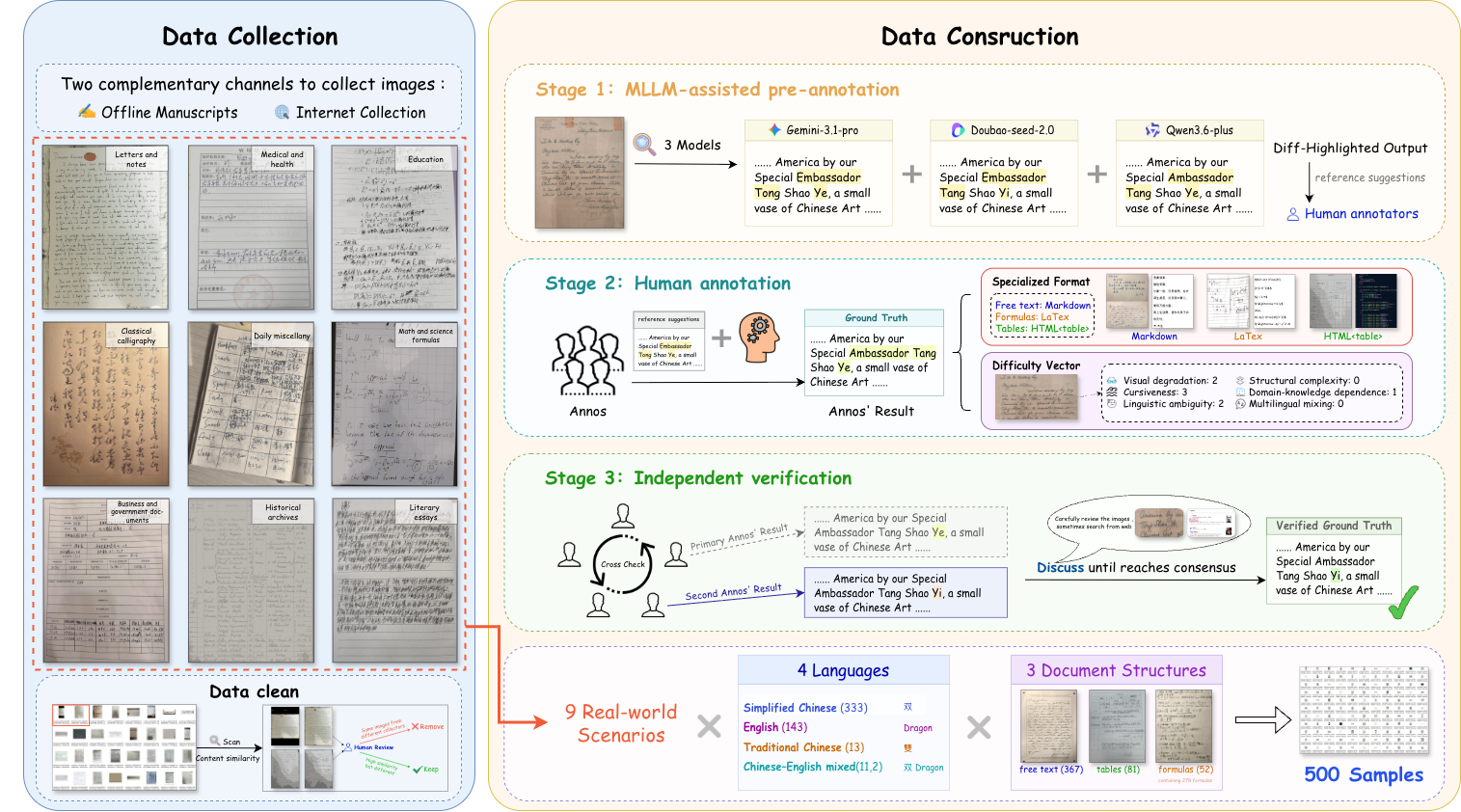}

\caption{
Overview of the \benchname{} construction pipeline.
}

\label{fig:pipeline}

\end{figure*}

\benchname{} jointly evaluates handwritten free text, tables, and formulas under realistic handwritten document scenarios. The benchmark contains 500 handwritten document images spanning three document structures, four language settings, and nine representative real-world scenarios. Figure~\ref{fig:pipeline} illustrates the overall construction pipeline.


\subsection{Benchmark Taxonomy}

To comprehensively characterize handwritten document understanding, every sample is organized along three dimensions.

\begin{itemize}

\item \textbf{Language (4).}
\benchname{} covers four language settings, including Simplified Chinese (333, 66.6\%), English (143, 28.6\%), Traditional Chinese (13, 2.6\%), and Chinese--English mixed documents (11, 2.2\%).

\item \textbf{Document Structure (3).}
The benchmark jointly evaluates three representative handwritten document structures:
free text (367, 73.4\%),
tables (81, 16.2\%),
and formulas (52 images containing 278 individually annotated formula regions, 10.4\%).

\item \textbf{Real-world Scenario (9).}
Samples are collected from nine representative scenarios, including literary writing, letters and notes, medical records, business and government documents, education, mathematical and scientific materials, classical calligraphy, daily miscellany, and historical archives.

\end{itemize}


\subsection{Benchmark Construction}

To maximize both dataset diversity and annotation reliability, handwritten documents are collected from two complementary sources.

\textbf{Offline handwritten manuscripts.}
Participants voluntarily contribute authentic handwritten materials covering multiple real-world domains. Since writers have direct access to the original content, reliable ground-truth transcriptions can be obtained from the source documents.

\textbf{Internet collection.}
Additional handwritten documents are collected from publicly available online resources where community discussions or decipherment references provide useful contextual information for annotation.

After collection, all samples undergo automatic content-similarity scanning to remove duplicates, followed by manual inspection of ambiguous cases.

Benchmark construction then proceeds in three stages.

\paragraph{Stage 1: MLLM-assisted pre-annotation.}

Three state-of-the-art MLLMs independently transcribe every handwritten document.
These predictions are not used as ground truth,
but serve as reference suggestions for human annotators,
helping identify ambiguous regions and improving annotation efficiency.

\paragraph{Stage 2: Human annotation.}

A primary annotator produces the recognition ground truth for every sample,
represented using Markdown for free text,
\LaTeX{} for formulas,
and HTML \texttt{<table>} for handwritten tables.
Additionally, each sample is annotated with difficulty ratings along six dimensions (visual degradation, cursiveness, linguistic ambiguity, structural complexity, domain knowledge, and multilingual mixing) to support future fine-grained analysis.

During annotation,
annotators consult all available evidence,
including MLLM predictions,
writer-provided transcripts,
community discussions,
and domain experts whenever necessary.
Characters that remain impossible to identify after exhaustive verification are masked in the original image and excluded from evaluation.

\paragraph{Stage 3: Independent verification.}

Every annotation is independently reviewed by a second annotator who performs the verification without access to the primary annotator's reasoning.
Disagreements are resolved through discussion until consensus is reached.
Only samples passing independent verification are included in the final benchmark.
This multi-stage quality-control protocol---automatic deduplication, MLLM-assisted annotation, independent double review, and consensus-based adjudication---substantially reduces annotation inconsistency while ensuring reliable ground truth for subsequent evaluation.

\section{Evaluation Protocol}

Our evaluation protocol consists of three components. First, structure-specific recognition metrics measure transcription quality across free text, formulas, and tables. Second, the proposed \textbf{Prior-Driven Error (PDE)} metric quantifies whether recognition failures originate from language priors rather than visual evidence. Third, calibrated human baselines provide a reference point for characterizing how human and machine errors differ qualitatively.

\subsection{Recognition Evaluation}

Following OmniDocBench~\cite{omnidocbench}, different handwritten document structures are evaluated using structure-specific metrics.

\begin{itemize}

\item \textbf{Free text.}
Character-level recognition quality is measured using normalized Edit Distance (Edit$\downarrow$)~\cite{levenshtein}.

\item \textbf{Formulas.}
Formula recognition is evaluated using CDM (Character Detection Metric, CDM$\uparrow$), which renders both prediction and ground truth before matching character bounding boxes~\cite{cdm}.

\item \textbf{Tables.}
Table understanding is evaluated using TEDS (Tree Edit Distance Similarity, TEDS$\uparrow$), which jointly measures structural correctness and textual fidelity based on HTML representations~\cite{pubtabnet}.

\end{itemize}

To facilitate comparison across different document structures, we report an overall score following OmniDocBench~\cite{omnidocbench}:

\[
\text{Overall}
=
\frac{
(1-\text{Edit})\times100
+
\text{CDM}
+
\text{TEDS}
}{3}.
\]

\subsection{Prior-Driven Error}

Recognition accuracy alone cannot distinguish different types of recognition failures. Two models may achieve identical recognition accuracy while exhibiting very different behaviors: one may fail because it cannot correctly perceive handwritten content, whereas the other may generate visually unsupported yet linguistically plausible transcriptions due to excessive reliance on language priors. Since these two failure modes require different modeling improvements, \benchname{} explicitly quantifies prior-driven recognition errors through the proposed \textbf{Prior-Driven Error (PDE)} metric.

For every prediction, model output is aligned with the corresponding ground truth using character-level alignment for free text and formulas, and cell-level alignment for tables. Only segments where prediction differs from the ground truth are considered. Each mismatched segment is then scored using an external reference language model (Qwen3-8B-Base~\cite{qwen3}). If the model-generated segment has lower perplexity than the corresponding ground-truth segment, the error is regarded as \emph{prior-driven}, indicating that the model replaces visually correct but less common content with a linguistically more probable alternative. Inserted content that has no corresponding ground truth is always regarded as prior-driven since it is generated solely from internal language priors.

Formally, for a sample $s$, the Prior-Driven Error rate is computed as

\[
\text{PDE}(s) = \frac{\sum_i \mathbb{1}[\text{ppl}(m_i ) < \text{ppl}(g_i)] \, |m_i|}{\sum_i |m_i|},
\]

where $m_i$ and $g_i$ denote the model prediction and the corresponding ground-truth segment, respectively. For inserted content, $\text{ppl}(g_i)=\infty$. The summation is performed only over mismatched segments. Consequently, PDE measures the proportion of model errors attributable to language priors rather than visual perception, instead of reflecting overall recognition accuracy. The overall PDE score reported for each model is the arithmetic mean of the per-category PDE rates (text, table, formula).


\subsection{Human Baselines}

\benchname{} additionally establishes calibrated human baselines under the same evaluation protocol. Human participants receive exactly the same handwritten document images as the evaluated models and independently produce recognition outputs using the same target formats. Performance is evaluated using the identical metrics described above.

To ensure a fair comparison, human participants perform single-pass recognition without access to any auxiliary resources. In particular, they are not allowed to consult MLLM predictions, writer-provided transcripts, community discussions, or domain experts. This setting differs intentionally from the benchmark construction process, where multiple evidence sources are integrated to establish reliable ground truth through consensus-based annotation.

Under this protocol, human participants achieve an overall score of 77.09\%, which is higher than the best MLLM (Gemini 3.1 Pro, 71.85\%). Detailed per-category comparisons are presented in the Experiments section.


\section{Experiments}
\label{sec:experiments}

\subsection{Evaluated Models}

\begin{figure*}[t]
\centering
\includegraphics[width=\textwidth]{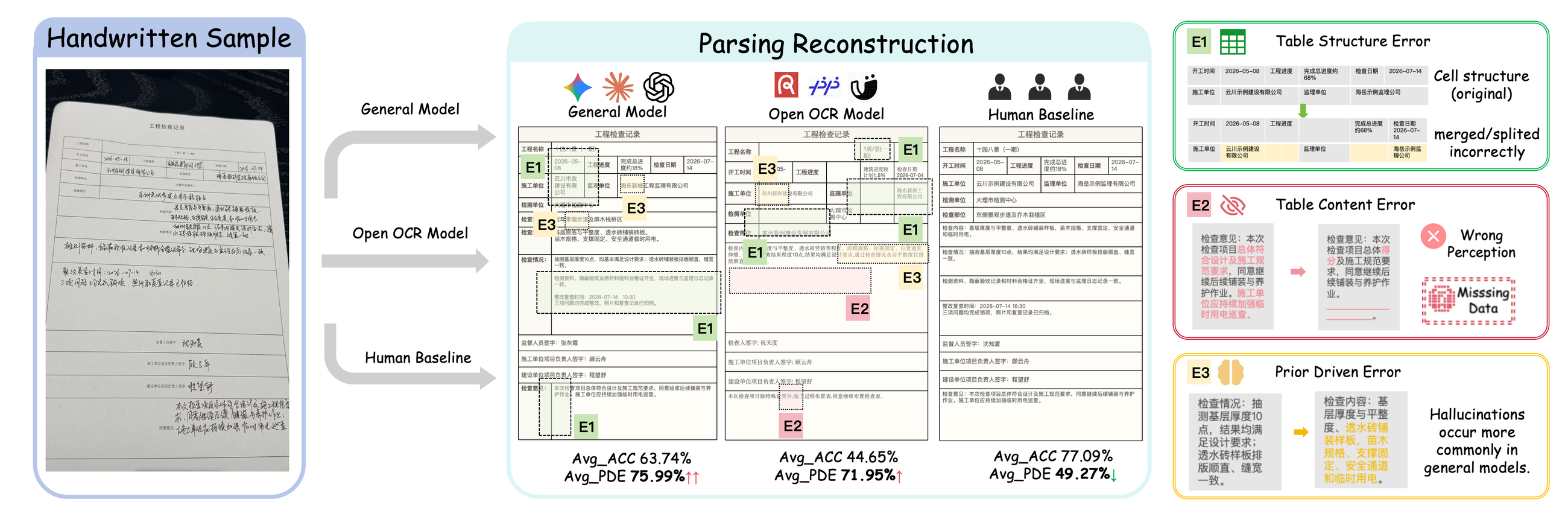}
\caption{Qualitative comparison of handwritten table parsing across general-purpose models, OCR-specialized models, and the human baseline. The right panel illustrates three error types: table structure errors (E1), table content errors (E2), and prior-driven errors (E3).}
\label{fig:error-pattern-analysis}
\end{figure*}

\begin{table*}[!ht]
\centering
\small
\setlength{\tabcolsep}{4.5pt}
\begin{tabular}{@{}ll c cccc c cccc@{}}
\toprule
 &  &  & \multicolumn{4}{c}{\textbf{Performance Evaluation (\%)}} & \phantom{a} & \multicolumn{4}{c}{\textbf{Prior-Driven Error (\%)}} \\
\cmidrule{4-7} \cmidrule{9-12}
\textbf{Type} & \textbf{Model} & \textbf{Params} & \makecell{Text\\Edit$\downarrow$} & \makecell{Table\\TEDS$\uparrow$} & \makecell{Formula\\CDM$\uparrow$} & \makecell{Overall\\$\uparrow$} & & Text$\downarrow$ & Table$\downarrow$ & Formula$\downarrow$ & Overall$\downarrow$ \\
\midrule
\multirow{3}{*}{\makecell{Proprietary}}
 & \logo{claude}Claude Opus 4.8 & -- & 53.30 & 52.27 & 74.11 & 57.70 & & 72.78 & 59.32 & 91.66 & 74.59 \\
 & \logo{gemini}Gemini 3.5 Flash & -- & 27.10 & 53.63 & \uline{78.22} & \uline{68.25} & & 71.36 & 71.91 & 94.36 & 79.21 \\
 & \logo{gemini}Gemini 3.1 Pro & -- & \textbf{24.32} & \textbf{60.44} & \textbf{79.42} & \textbf{71.85} & & 71.19 & 66.56 & 94.77 & 77.51 \\
\midrule
\multirow{4}{*}{\makecell{Open General}}
 & \logo{intern}InternVL3.5 & 241B-A28B & 46.80 & 41.22 & 64.00 & 52.80 & & 72.62 & 66.38 & 90.03 & 76.34 \\
 & \logo{qwen}Qwen3-VL & 235B-A22B & 34.30 & 51.81 & 69.06 & 62.19 & & 70.01 & 62.18 & 91.33 & 74.51 \\
 & \logo{qwen}Qwen3.5-Plus & 397B-A17B & 34.80 & 57.68 & 73.77 & 65.55 & & 71.74 & 70.00 & 91.71 & 77.82 \\
 & \logo{kimi}Kimi-K3 & 2.8T-104B & \uline{26.07} & \uline{58.99} & 70.50 & 67.81 & & 65.29 & 60.32 & 90.21 & 71.94 \\
\midrule
\multirow{11}{*}{\makecell{Open OCR}}
 & \logo{deepseek}DeepSeek-OCR2 & 3B-A0.5B & 91.70 & 1.77 & 18.33 & 9.46 & & 84.06 & 90.00 & 98.35 & 90.80 \\
 & \logo{mineru}MinerU-2.5 & 1.2B & 83.80 & 41.19 & 36.56 & 31.33 & & 73.80 & 55.21 & 92.16 & 73.72 \\
 & \logo{baidu}Unlimited-OCR & 3B-A0.5B & 64.90 & 34.41 & 51.62 & 40.38 & & 61.67 & 55.16 & \uline{89.47} & 68.77 \\
 & \logo{xiaohongshu}Dots.ocr & 3B & 56.68 & 34.70 & 53.17 & 43.73 & & 70.29 & 59.75 & 90.98 & 73.67 \\
 & \logo{xiaohongshu}Dots.mocr & 3B & 49.54 & 36.60 & 62.85 & 49.97 & & 63.23 & 59.56 & 89.73 & 70.84 \\
 & \logo{hunyuan}HunyuanOCR & 1.0B & 38.50 & 47.02 & 45.81 & 51.44 & & 67.13 & 55.55 & 92.08 & 71.59 \\
 & \logo{mineru}MinerU2.5-Pro & 1.2B & 59.80 & 53.16 & 61.79 & 51.73 & & 61.23 & \uline{47.46} & 92.80 & 67.16 \\
 & \logo{baidu}Qianfan-OCR & 4B & 50.40 & 45.22 & 61.40 & 52.07 & & 64.07 & 61.29 & 94.04 & 73.13 \\
 & \logo{xiaohongshu}FireRed-OCR & 2B & 49.00 & 44.30 & 63.11 & 52.82 & & 64.70 & 64.25 & 93.10 & 74.02 \\
 & \logo{glmv}GLM-OCR & 0.9B & 54.20 & 57.38 & 59.01 & 54.06 & & \uline{57.02} & \textbf{41.92} & 90.89 & \textbf{63.28} \\
 & \logo{paddle}PaddleOCR-VL-1.6 & 0.9B & 41.52 & 52.41 & 51.75 & 54.21 & & \textbf{52.59} & 53.01 & \textbf{87.83} & \uline{64.48} \\
\midrule
Human & Human & -- & 30.08 & 74.78 & 86.56 & 77.09 & & 66.96 & 34.48 & 46.36 & 49.27 \\
\bottomrule
\end{tabular}
\caption{Recognition performance and prior-driven error for all evaluated models and the human baseline.}
\label{tab:performance}
\end{table*}

We evaluate 18 models grouped into three categories: proprietary VLMs (Claude Opus 4.8~\cite{claude_opus48}, Gemini 3.1 Pro~\cite{gemini31pro}, Gemini 3.5 Flash~\cite{gemini35flash}), open-source general VLMs (InternVL3.5~\cite{internvl35}, Qwen3-VL~\cite{qwen3vl}, Qwen3.5-Plus~\cite{qwen35plus}, Kimi-K3~\cite{kimi_k3}), and open-source OCR-focused VLMs (DeepSeek-OCR-V2~\cite{deepseek_ocr2}, MinerU2.5~\cite{mineru25}, MinerU2.5-Pro~\cite{mineru25pro}, Unlimited-OCR~\cite{unlimited_ocr}, Dots.ocr~\cite{dots_ocr}, Dots.mocr~\cite{dots_mocr}, HunyuanOCR~\cite{hunyuan_ocr}, Qianfan-OCR~\cite{qianfan_ocr}, FireRed-OCR~\cite{firered_ocr}, GLM-OCR~\cite{glm_ocr}, PaddleOCR-VL-1.6~\cite{paddleocr_vl16}).

All models are evaluated with an identical post-processing pipeline. General VLMs use the standard OmniDocBench prompt template~\cite{omnidocbench}, while OCR-focused models use their respective recommended prompt settings or APIs.

Models are selected to represent the current state of the art in document understanding, with reference to the OmniDocBench leaderboard~\cite{omnidocbench}.

\subsection{Main Results}

Table~\ref{tab:performance} presents the main results. The most striking finding is the performance degradation from printed to handwritten documents: models that exceed 90\% on OmniDocBench~\cite{omnidocbench} drop to 71.85\% at best on \benchname{}, confirming a persistent ``handwritten gap.'' Furthermore, OCR-focused models---which routinely outperform general VLMs on printed-document benchmarks---fall behind general VLMs on wild handwriting. The top six models on \benchname{} are all general-purpose, suggesting that larger model capacity and more diverse training data enable stronger generalization to the high variability of handwritten content, while OCR-focused models are specialized for the regularity of printed layouts.

Tables remain the most challenging category: even the best TEDS score (60.44) indicates nearly 40\% structural and content mismatch, and table scores show the smallest cross-model variance among functional models.
Text recognition shows the widest spread, reflecting large differences in character-level fidelity.
Gemini 3.1 Pro leads in all three categories and overall, though Kimi-K3 is competitive on text edit distance (26.07 vs.\ 24.32) and Gemini 3.5 Flash on formula CDM (78.22 vs.\ 79.42).

\subsection{Human vs.\ Model Analysis}

The human baseline achieves an overall score of 77.09\%---\emph{higher} than the best MLLM (Gemini 3.1 Pro, 71.85\%) by 5.24 absolute points.
While this confirms that human performance remains the upper bound, the relatively modest gap indicates that wild handwriting challenges humans and models alike.

The per-category breakdown reveals where models still lag behind humans and where they have caught up.
On text, the best model actually surpasses humans (Edit: 30.08 vs.\ 24.32), indicating that MLLMs produce more format-compliant transcriptions even though humans may achieve superior character-level recognition.
On tables (TEDS: 74.78 vs.\ 60.44), humans substantially outperform all models by over 14 absolute points, reflecting superior structural understanding of irregular handwritten tables.
On formulas (CDM: 86.56 vs.\ 79.42), humans also lead, though the gap is smaller (7.14 points), indicating that formula recognition in current MLLMs is relatively mature.

Crucially, human errors are qualitatively different from model errors. The PDE analysis shows that human PDE is 49.27\%, substantially lower than all evaluated models (minimum 63.28\%), indicating that human errors are more evenly split between visual misrecognition and prior-driven substitution. Models, in contrast, produce disproportionately prior-driven errors---generating fluent but visually unsupported output where humans tend toward conservative partial transcriptions. This qualitative gap in failure modes persists despite the narrowing quantitative gap in overall accuracy.

\subsection{Error Pattern Analysis}

Table~\ref{tab:performance} also reports the prior-driven error rate decomposed by content structure; Figure~\ref{fig:error-pattern-analysis} provides qualitative examples of table parsing errors across model types and the human baseline.
Among models with functional visual encoders, PDE rates range from 63\% to 79\% without a simple correlation to accuracy. Notably, general VLMs with strong language capabilities show relatively high rates (Gemini 3.5 Flash: 79.21\%, Qwen3.5-Plus: 77.82\%, Gemini 3.1 Pro: 77.51\%), while the lowest PDE rates belong to certain compact OCR-focused models (GLM-OCR: 63.28\%, PaddleOCR-VL-1.6: 64.48\%).

This reveals a nuanced relationship between model capability and error type.
At the extreme, models with severely limited visual decoding produce almost exclusively prior-driven errors.
Among better-performing models, powerful language priors act as a double-edged sword: they boost accuracy by correctly inferring ambiguous characters, but when errors do occur, those errors are disproportionately prior-driven---the model's strong language capability causes it to confidently generate plausible alternatives rather than fail silently.
Conversely, the models with the lowest PDE rates (GLM-OCR, PaddleOCR-VL) are compact models that produce a larger proportion of visual misrecognition errors (e.g., confusing similar-looking characters) rather than generating fluent but unsupported text.

The per-structure breakdown reveals that formula errors are almost entirely prior-driven across all models (87--98\%), likely because LaTeX syntax is inherently low-perplexity---even legitimate format differences satisfy the ppl criterion.
Text errors show the widest spread (52--84\%), reflecting the greatest variation in model reliance on language priors.
Table errors are moderate (42--90\%), with weaker models showing higher rates.

We note limitations of this metric: it characterizes error \emph{type}, not error \emph{quantity}---a high prior-driven rate does not mean a model is less reliable, only that its errors are more systematic.
The classification depends on a specific reference language model (Qwen3-8B-Base); a different model may shift boundary cases.

\section{Discussion}

The findings of \benchname{} demonstrate that handwritten document understanding remains unsolved: models exceeding 90\% on OmniDocBench~\cite{omnidocbench} drop to 71.85\% at best on \benchname{}, and this degradation is non-uniform across document structures. More critically, the PDE analysis reveals that model and human errors are qualitatively different---63--79\% of model errors among functional models originate from language priors, compared to 49\% for humans. This means that model errors are systematically more dangerous: fluent, confident, and unsupported by visual evidence.

These findings have implications for safety-critical applications such as medical records, financial documents, and legal archives, where prior-driven errors are particularly difficult to detect~\cite{hallucination_survey}. Future models should optimize not only recognition accuracy but also visual grounding and calibrated uncertainty, learning to express uncertainty rather than confidently generating plausible transcriptions when visual evidence is insufficient.

\subsection{Limitations}

\textbf{Benchmark scale and coverage.}
Although \benchname{} contains carefully curated handwritten documents with high-quality annotations, it currently consists of 500 samples and should not be interpreted as a population-level estimate of handwritten OCR performance. The relatively small formula (52 images) and table (81 images) subsets limit statistical power for distinguishing highly competitive models. Furthermore, the benchmark focuses primarily on Chinese and English handwriting. Future versions will expand both language coverage and dataset scale.

\textbf{Annotation and human evaluation.}
Part of the benchmark is collected from publicly available Internet resources, which may introduce source bias despite careful manual screening and deduplication. Certain handwritten scenarios, such as historical calligraphy or severely degraded manuscripts, inherently involve subjective interpretation, making a universally correct transcription difficult to define even under consensus-based annotation. The reported human baseline should therefore be regarded as a calibrated reference rather than an absolute upper bound. Moreover, all participating annotators are native Chinese speakers, which may underestimate achievable human performance on highly cursive or stylistically diverse English handwriting.

\textbf{PDE metric.}
The PDE metric depends on an external reference language model (Qwen3-8B-Base~\cite{qwen3}) for perplexity estimation, and alternative reference models may lead to different classifications for ambiguous boundary cases. Moreover, PDE is designed to characterize error \emph{type} rather than quantity---a higher PDE does not imply lower model reliability, only that a larger proportion of errors originate from language priors rather than visual perception.

\section{Conclusion}

We presented \benchname{}, a benchmark for handwritten document understanding that jointly evaluates free text, tables, and formulas with a unified protocol combining recognition metrics, human baselines, and Prior-Driven Error analysis. Experiments on 18 models reveal a persistent handwritten gap (best model 71.85\% vs.\ human 77.09\%), with model errors qualitatively different from human errors---63--79\% prior-driven for models vs.\ 49\% for humans.

\bibliography{references}

\clearpage
\onecolumn
\appendix
\section*{Supplementary Material}

\noindent
Section~A provides additional dataset construction details beyond those in the main paper.
Section~B presents qualitative PDE (Prior-Driven Error) examples from \benchname{}.
Section~C gives complete evaluation details including the full model list, prompt templates,
and inference configuration.

\section*{A. Dataset Details}

This section expands on the dataset construction summarized in the main paper. We report only
details not already covered there, including per-source sample counts, the joint
category--language distribution, per-scenario counts, and the annotation and image-processing
specifications.

\subsection*{A.1\quad Data Collection Sources}

The 500 samples are drawn from two complementary channels:
\begin{itemize}
\item \textbf{Offline handwritten manuscripts (141 samples, 28.2\%).}
Voluntary contributions from project members and acquaintances, concentrated in the education
and medical domains (classroom notes, homework, medical records, prescriptions), plus
hand-transcribed article abstracts, reading notes, and letters to broaden writing styles.
\item \textbf{Internet collection (359 samples, 71.8\%).}
Publicly available handwritten images covering long-tail genres: classical poetry and calligraphy,
historical archival registers, old newspapers, and official forms.
\end{itemize}

\subsection*{A.2\quad Category $\times$ Language Distribution}

The main paper reports the marginal counts for document structure and language separately.
Table~\ref{tab:cat_lang} gives the full joint distribution.

\begin{table}[h]
\centering
\caption{Category $\times$ Language joint distribution.}
\label{tab:cat_lang}
\begin{tabular}{lrrrr|r}
\toprule
Category & zh\_hans & en & zh\_hant & zh\_en\_mix & Total \\
\midrule
Text    & 251 & 99 & 12 & 7  & 369 \\
Table   & 62  & 16 & 0  & 1  & 79  \\
Formula & 20  & 29 & 0  & 3  & 52  \\
\midrule
Total   & 333 & 144 & 12 & 11 & 500 \\
\bottomrule
\end{tabular}
\end{table}

\subsection*{A.3\quad Scenario Distribution}

The main paper enumerates the nine scenarios; Table~\ref{tab:scenario} adds their per-scenario
sample counts.

\begin{table}[h]
\centering
\caption{Per-scenario sample counts.}
\label{tab:scenario}
\begin{tabular}{lr}
\toprule
Scenario & Count \\
\midrule
Literary essays           & 93 \\
Letters \& notes          & 81 \\
Medical \& health         & 62 \\
Education \& learning     & 61 \\
Business documents        & 58 \\
Math \& science formulas  & 55 \\
Classical poetry \& calligraphy & 41 \\
Daily life notes          & 32 \\
Historical archives       & 17 \\
\midrule
Total & 500 \\
\bottomrule
\end{tabular}
\end{table}

\subsection*{A.4\quad Difficulty Annotation Protocol}

The six difficulty dimensions defined in the main paper are each scored independently by
5 reviewers per sample, and per-dimension scores
are averaged across reviewers. These ratings support fine-grained analysis but are not used in
the aggregate benchmark scores.

\subsection*{A.5\quad Annotation Format Details}

Beyond the target formats stated in the main paper (Markdown for free text, \LaTeX{} for formulas,
HTML \texttt{<table>} for tables), formula regions are additionally annotated with \emph{polygon}
coordinates (closed polygons rather than axis-aligned boxes) to accommodate slanted handwritten
layouts, with each polygon paired one-to-one with its \LaTeX{} transcription. Table annotations
carry full structural markup (\texttt{rowspan}, \texttt{colspan}, header hierarchy).

\subsection*{A.6\quad Image Processing}

All images are stored in PNG format with the following specifications:
\begin{itemize}
\item Maximum file size: 1\,MB (palette optimization applied where needed).
\item EXIF normalization: rotation tags are applied and stripped.
\item Naming convention: \texttt{hardhand\_<UUID\_no\_dashes>.png} (32-character hex UUID).
\end{itemize}

\section*{B. Qualitative PDE Examples}

\begin{figure}[h]
\centering
\includegraphics[width=\linewidth]{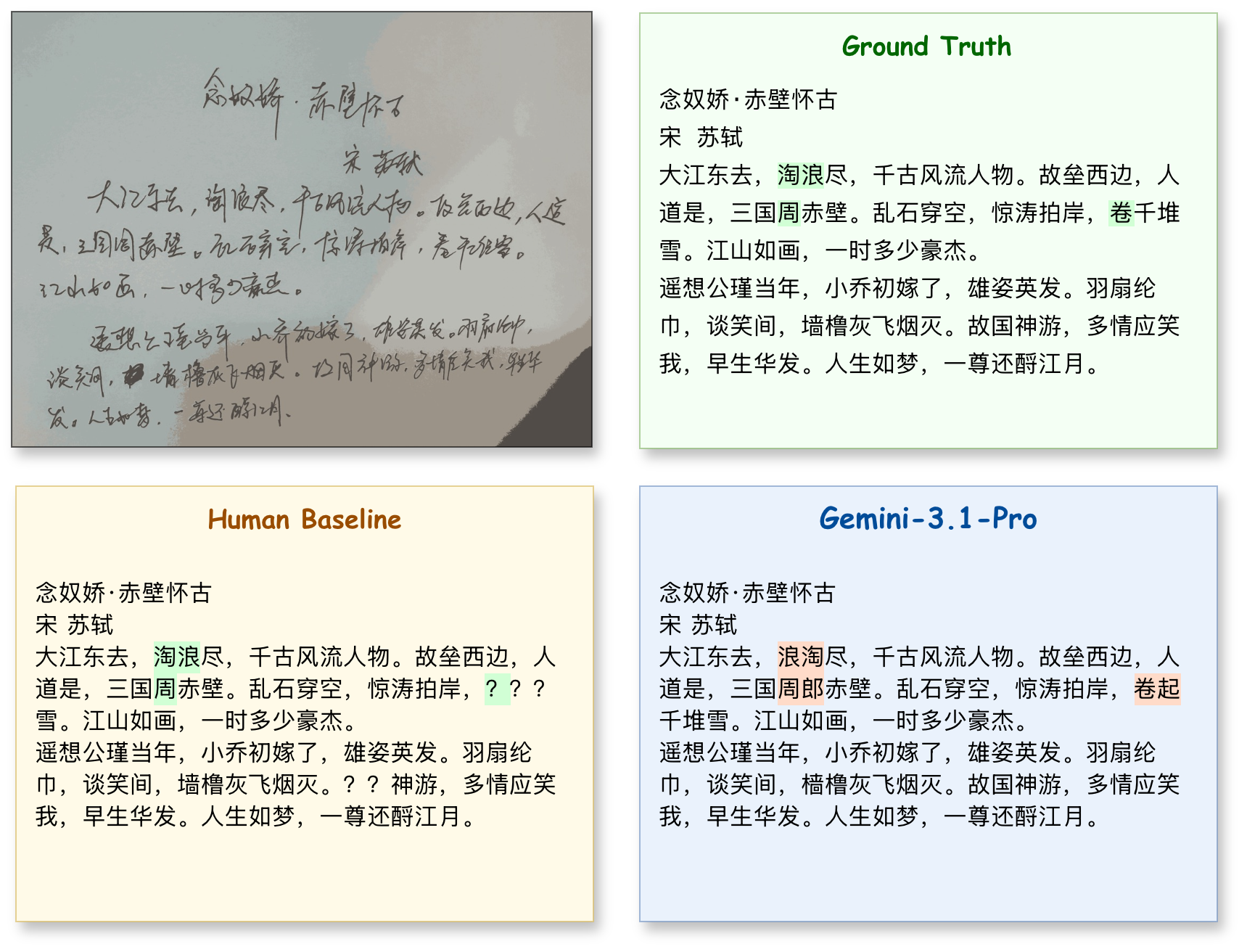}
\caption{Classical poem (calligraphy). A handwritten copy of Su Shi's \emph{Chibi Nostalgia} contains
non-standard character orderings. Gemini ``corrects'' the handwriting to
the memorized textbook version---swapping the non-standard two-character ordering (\emph{t\'ao~l\`ang})
back to the canonical order (\emph{l\`ang~t\'ao}).
This is a classic canonical-text prior: the model overwrites the actual handwritten content with a
remembered standard version. The human baseline faithfully preserves the non-standard forms and marks
only truly illegible characters with ``?''.}
\label{fig:pde_poem}
\end{figure}

\begin{figure}[h]
\centering
\includegraphics[width=\linewidth]{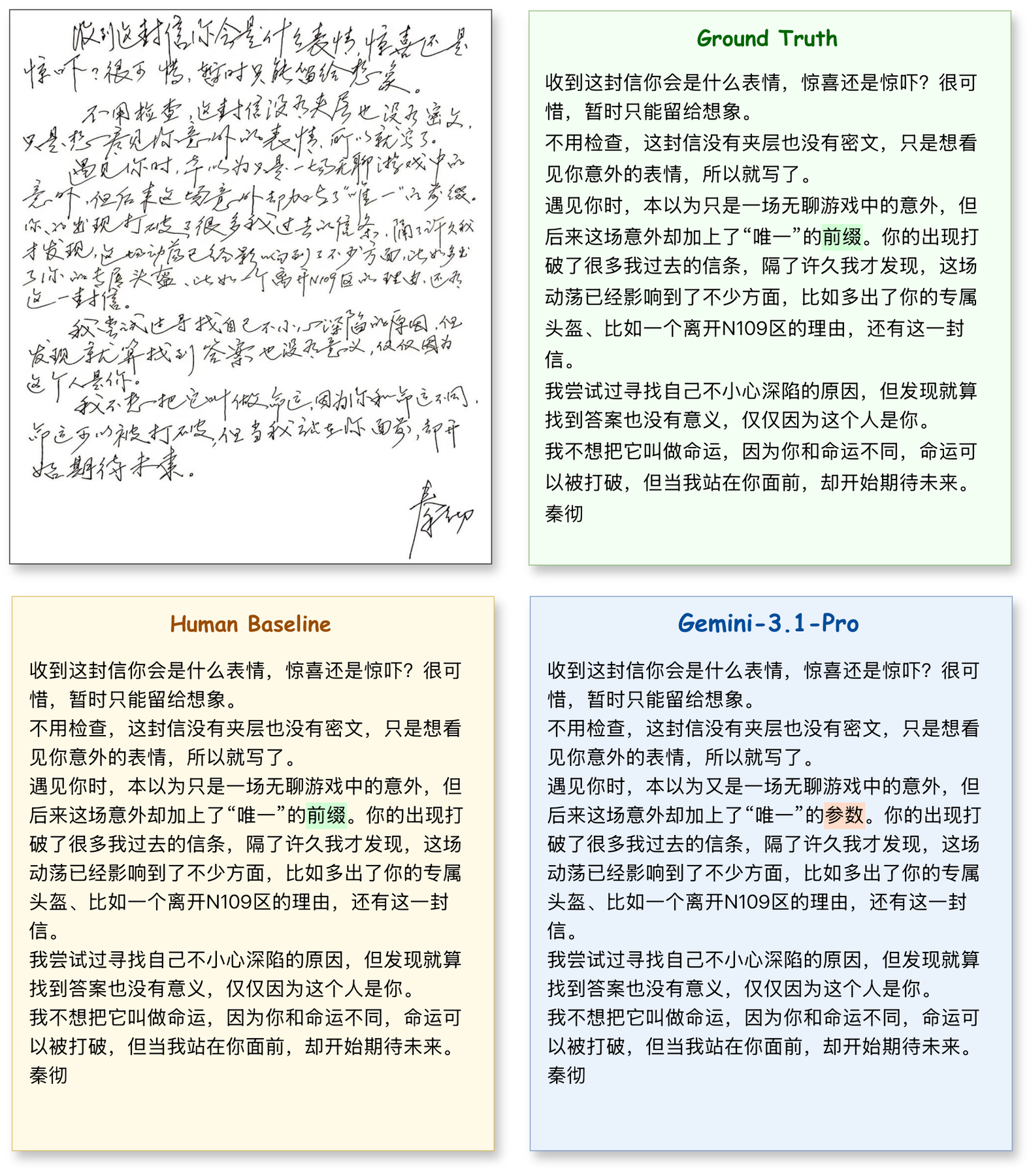}
\caption{Personal letter. Gemini substitutes ``prefix'' with ``parameter''---a semantic-field
substitution in which the model replaces the written word with a semantically related term drawn
from its priors rather than transcribing the source faithfully.}
\label{fig:pde_letter}
\end{figure}

\begin{figure}[h]
\centering
\includegraphics[width=\linewidth]{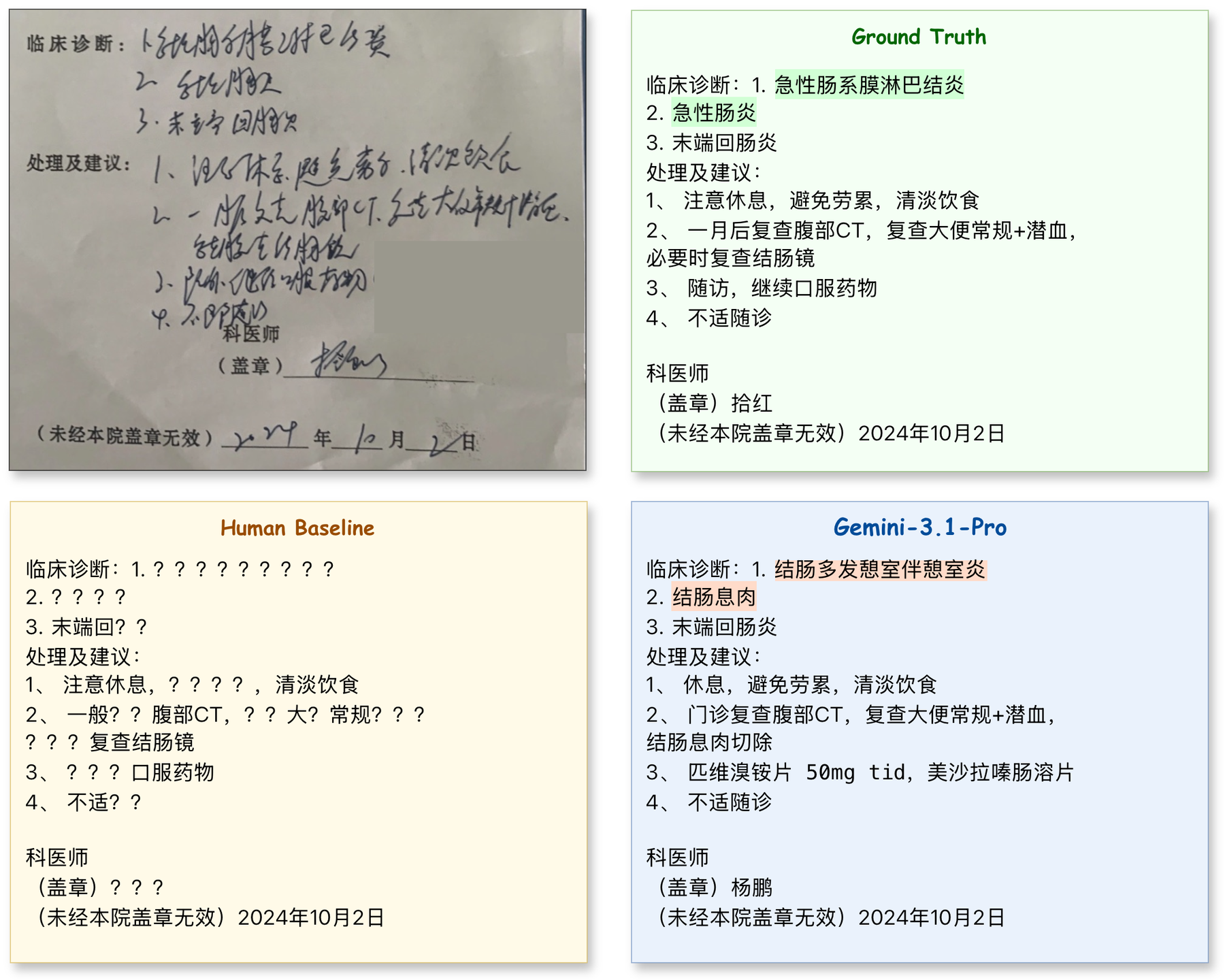}
\caption{Clinical diagnosis. The GT diagnosis is ``acute mesenteric lymphadenitis / acute enteritis /
terminal ileitis.'' Gemini outputs a completely different yet medically plausible diagnosis
(``colonic diverticulitis / colonic polyps / terminal ileitis''; only the third item matches). This is
the most extreme form of PDE: medical domain-knowledge priors fabricate a coherent but factually wrong
diagnosis. The human baseline honestly writes ``?'' for illegible portions.}
\label{fig:pde_medical}
\end{figure}

\begin{figure}[h]
\centering
\includegraphics[width=\linewidth]{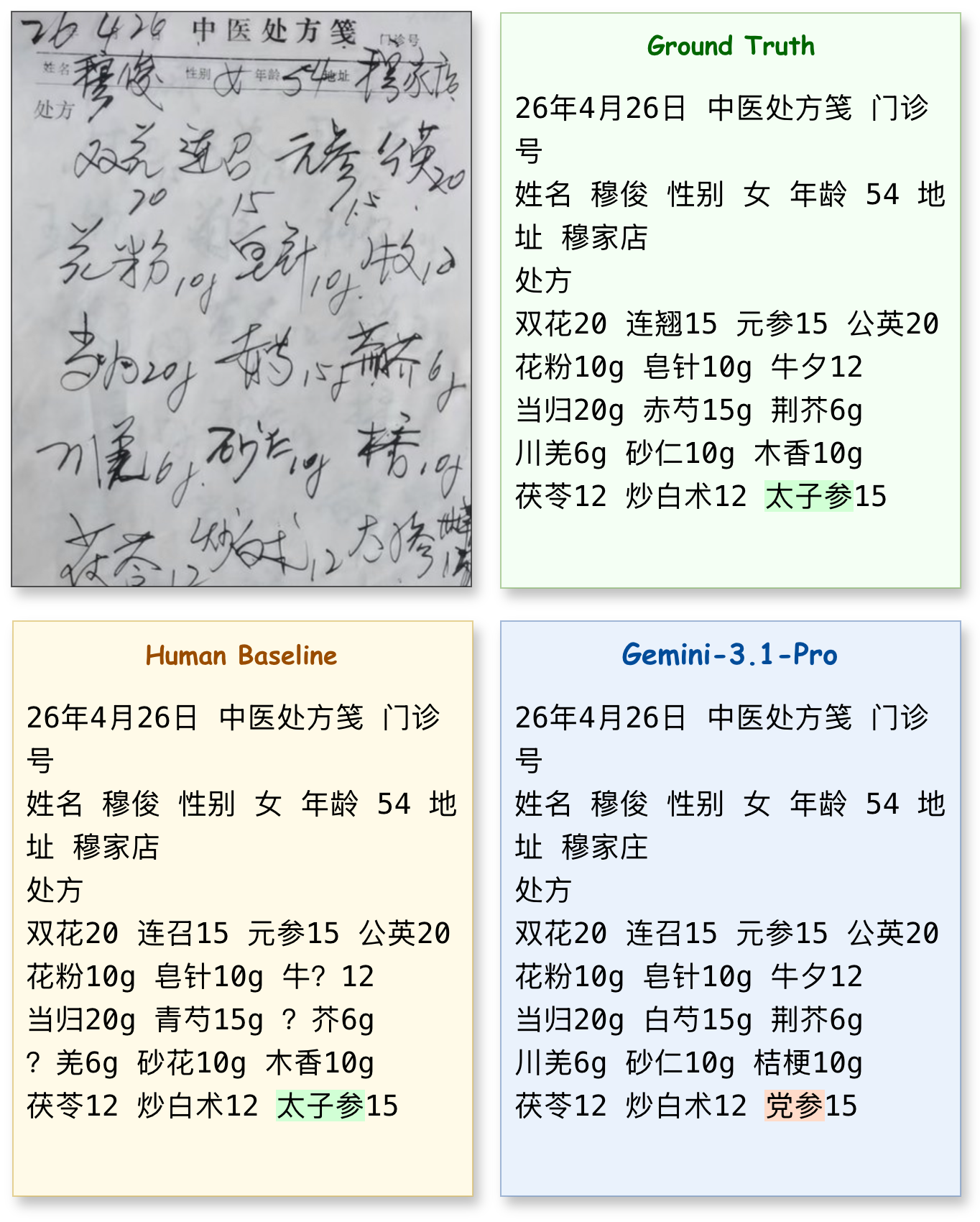}
\caption{Traditional Chinese medicine prescription. Gemini applies a prior-driven substitution,
replacing a herb name with a more frequently prescribed alternative.}
\label{fig:pde_prescription}
\end{figure}

\begin{figure}[h]
\centering
\includegraphics[width=\linewidth]{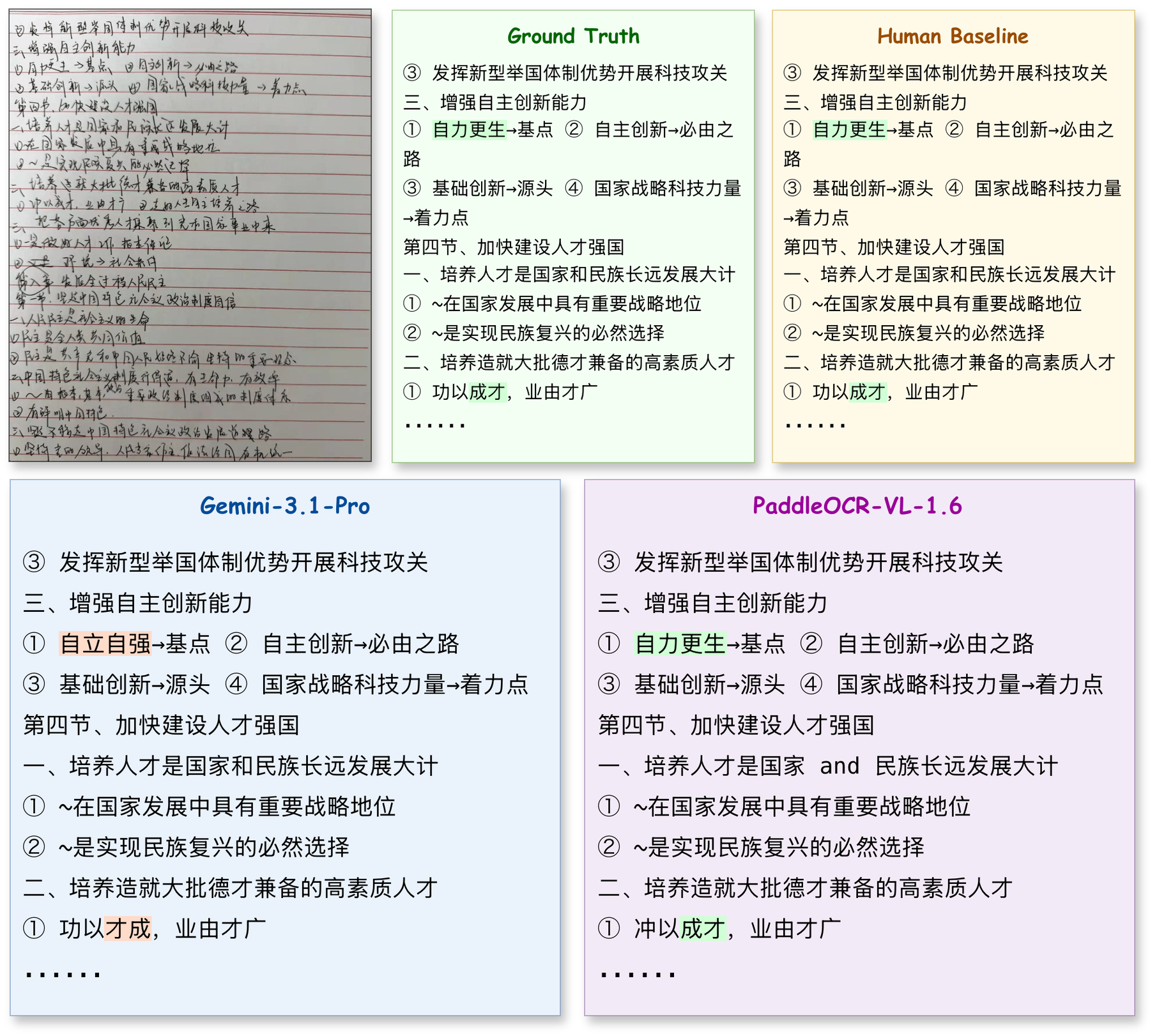}
\caption{Political study notes. Gemini rewrites into more formal policy terminology
(``self-reliance'' $\to$ ``self-strengthening'') and swaps the character order inside a
four-character phrase (``achievement makes talent'' $\to$ ``talent makes achievement'')---both
linguistically valid but not matching the source. This cleanly
contrasts high PDE (semantic rewriting) against low PDE (visual misreading).}
\label{fig:pde_notes}
\end{figure}

\begin{figure}[h]
\centering
\includegraphics[width=\linewidth]{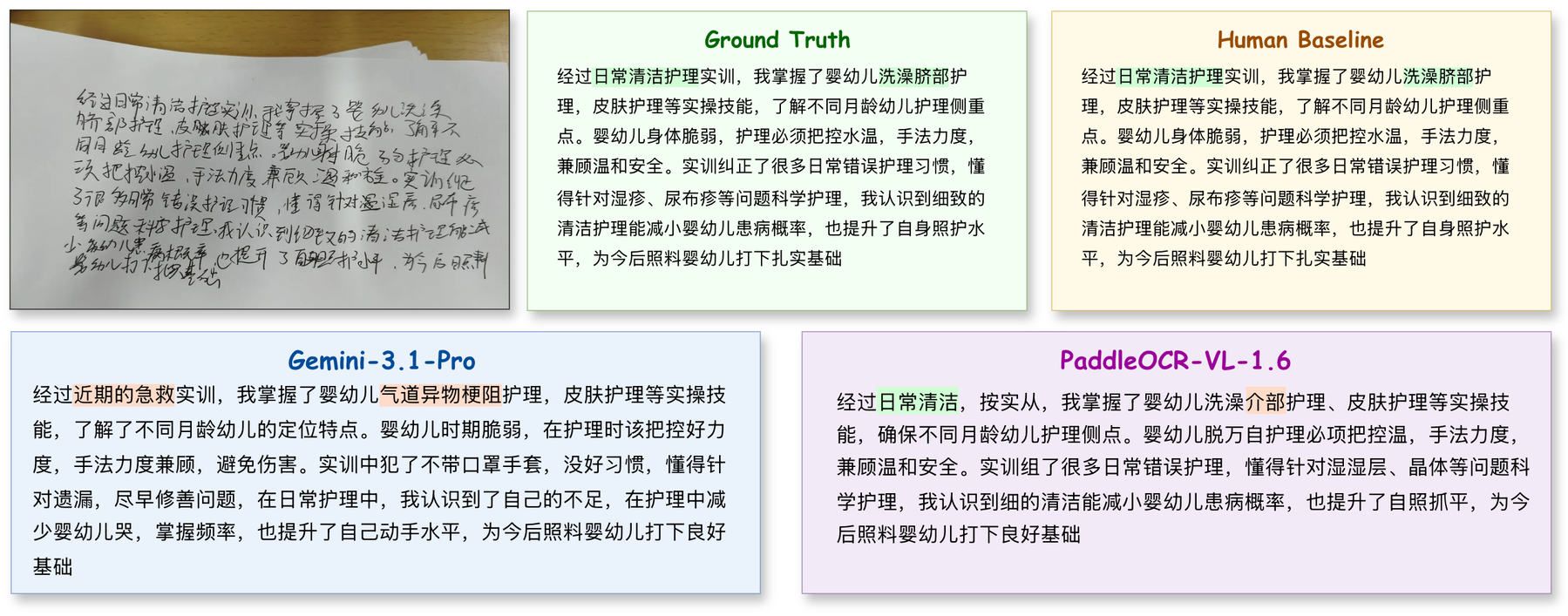}
\caption{Childcare training reflection. Gemini fabricates an entirely different topic: ``daily cleaning
skills training'' becomes ``recent first-aid training,'' and ``infant bathing / umbilical care'' becomes
``infant airway obstruction care''---the whole passage topic is replaced. PaddleOCR-VL exhibits
character-level noise but preserves the correct topic keywords (``daily cleaning,'' ``bathing'').
The human baseline matches the GT exactly.}
\label{fig:pde_reflection}
\end{figure}

\begin{figure}[h]
\centering
\includegraphics[width=\linewidth]{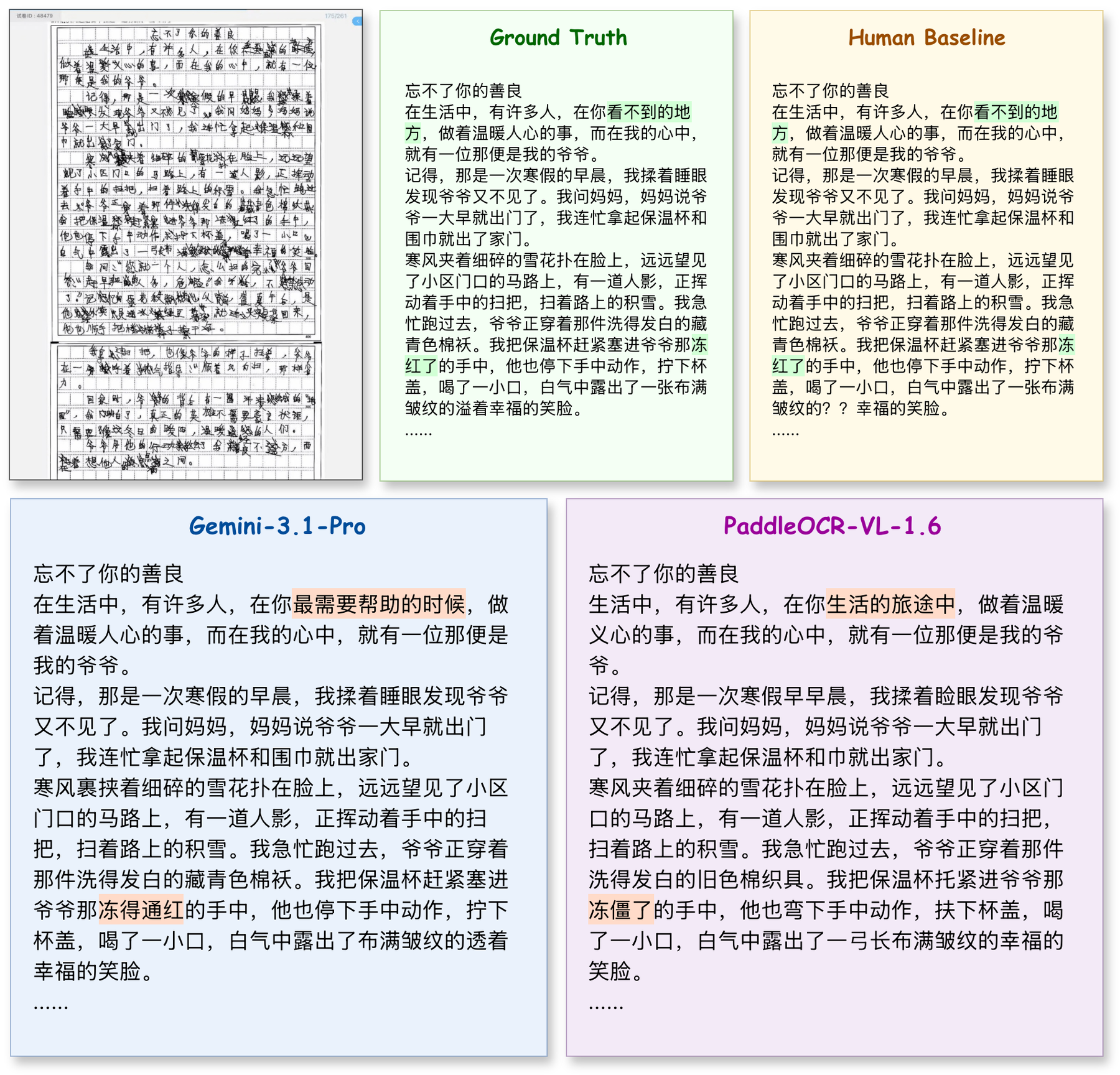}
\caption{Elementary-school essay. The student wrote ``doing heart-warming things \emph{in places you
cannot see}.'' Gemini polishes it to ``doing heart-warming things \emph{when you most need help}''---%
semantically ``better'' but not the original text (PDE). Both Gemini and PaddleOCR-VL are effectively
``polishing'' the child's composition.}
\label{fig:pde_essay}
\end{figure}

\begin{figure}[h]
\centering
\includegraphics[width=\linewidth]{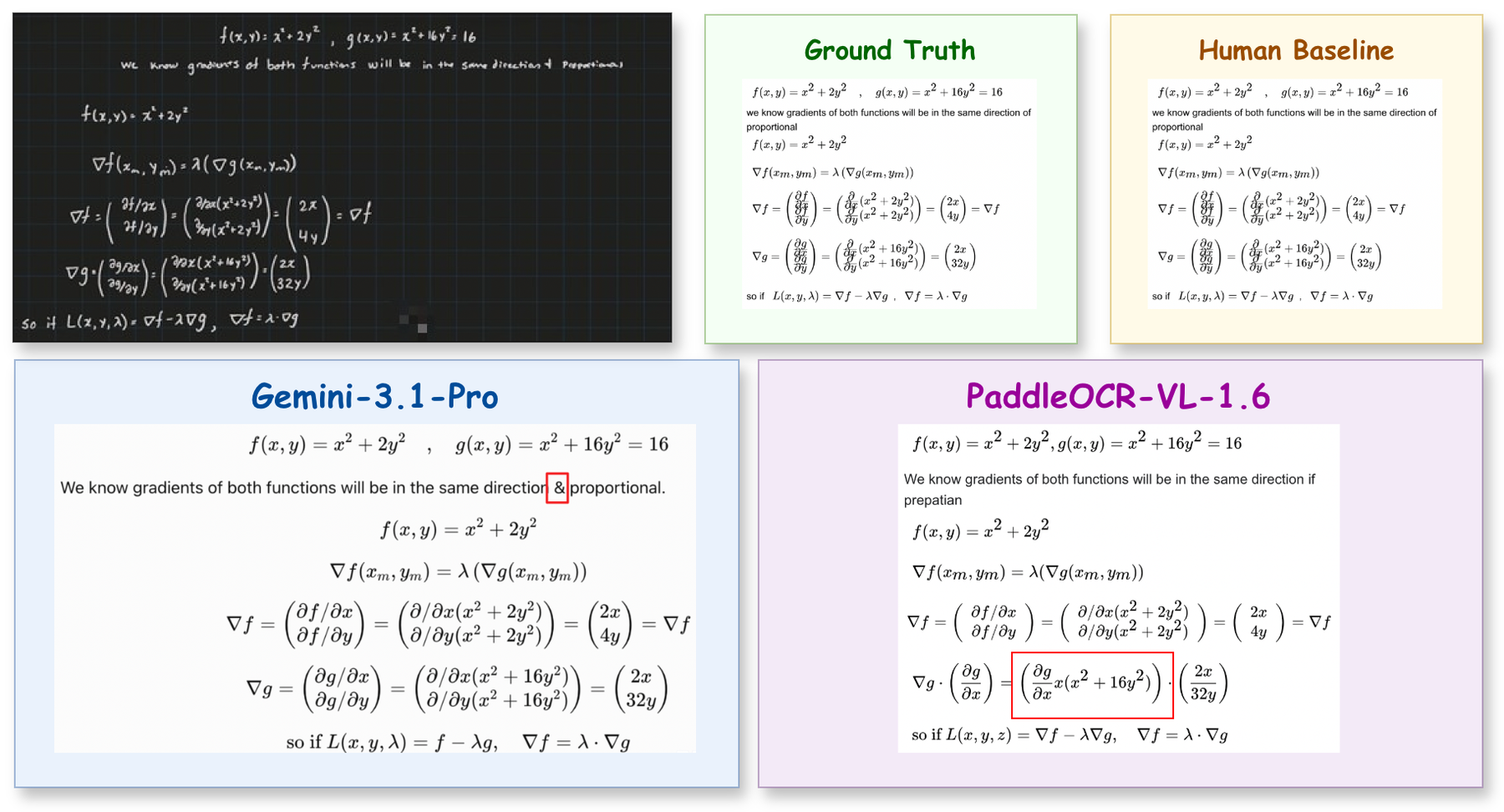}
\caption{Mathematical derivation (Lagrange multipliers). Gemini corrects the awkward phrase
``in the same direction of proportional'' to ``in the same direction \& proportional.''
PaddleOCR-VL misreads ``proportional'' as ``prepatian'' and garbles the $\nabla g$ matrix line,
but shows no semantic rewriting.}
\label{fig:pde_lagrange}
\end{figure}

\clearpage
\section*{C. Complete Evaluation Details}

\subsection*{C.1\quad Model List \& Access Details}

Table~\ref{tab:models} lists all evaluated models with their access method and model identifier.

\begin{table}[t]
\centering
\caption{Complete list of evaluated models.}
\label{tab:models}
\small
\begin{tabular}{llll}
\toprule
Model Name & Type & Access & Model ID / Version \\
\midrule
GPT-5.4              & General VLM   & API (OpenAI-compat) & \texttt{gpt-5.4} \\
Gemini-3.1-Pro       & General VLM   & API (OpenAI-compat) & \texttt{gemini-3.1-pro-preview} \\
Gemini-3.5-Flash     & General VLM   & API (OpenAI-compat) & \texttt{gemini-3.5-flash} \\
Claude-Opus-4-8      & General VLM   & API (OpenAI-compat) & \texttt{claude-opus-4-8} \\
Qwen3.5-Plus         & General VLM   & API (OpenAI-compat) & \texttt{qwen3.5-plus} \\
Qwen3-VL             & General VLM   & API (vLLM)          & \texttt{qwen3-vl-235b-a22b-thinking} \\
InternVL3.5          & General VLM   & API                 & \texttt{internvl3.5-241b-a28b} \\
Kimi-K3              & General VLM   & API (Anthropic fmt)  & \texttt{kimi-k3} \\
\midrule
Qianfan-OCR          & OCR-focused   & API (Baidu)         & \texttt{qianfan-ocr} \\
Hunyuan-OCR          & OCR-focused   & Local (vLLM)        & \texttt{tencent/HunyuanOCR} \\
GLM-OCR              & OCR-focused   & API (ZhipuAI)       & \texttt{glm-ocr} \\
PaddleOCR-VL         & OCR-focused   & API (async job)     & \texttt{PaddleOCR-VL-1.6} \\
DeepSeek-OCR-v2      & OCR-focused   & Local (vLLM)        & \texttt{deepseek-ai/DeepSeek-OCR} \\
FireRed-OCR          & OCR-focused   & Local (vLLM)        & \texttt{FireRed-OCR} \\
Unlimited-OCR        & OCR-focused   & Local (vLLM)        & \texttt{Unlimited-OCR} \\
\midrule
MinerU-2.5           & Document parser & Local (vLLM)      & \texttt{opendatalab/MinerU2.5-2509-1.2B} \\
MinerU-2.5-Pro       & Document parser & Local (vLLM)      & \texttt{opendatalab/MinerU2.5-Pro} \\
\bottomrule
\end{tabular}
\end{table}

\subsection*{C.2\quad Prompt Templates}

All general-purpose VLMs (GPT-5.4, Gemini-3.1-Pro, Gemini-3.5-Flash, Claude-Opus-4-8, Qwen3.5-Plus,
Qwen3-VL, InternVL3.5, Kimi-K3) use the standard OmniDocBench prompt. For all OCR-focused models
(Qianfan-OCR, Hunyuan-OCR, GLM-OCR, PaddleOCR-VL, DeepSeek-OCR-v2, FireRed-OCR, Unlimited-OCR)
and document parsers (MinerU-2.5, MinerU-2.5-Pro), we use the default API parameters when calling
their APIs, or the default settings of the model when deployed locally---no user-supplied prompt is
overridden.

\paragraph{OmniDocBench prompt (default).}

\begin{lstlisting}
You are an AI assistant specialized in converting PDF
images to Markdown format. Please follow these
instructions for the conversion:

1. Text Processing:
- Accurately recognize all text content in the PDF
  image without guessing or inferring.
- Convert the recognized text into Markdown format.
- Maintain the original document structure, including
  headings, paragraphs, lists, etc.

2. Mathematical Formula Processing:
- Convert all mathematical formulas to LaTeX format.
- Enclose inline formulas with \( \). For example:
  This is an inline formula \( E = mc^2 \)
- Enclose block formulas with \[ \]. For example:
  \[ \frac{-b \pm \sqrt{b^2 - 4ac}}{2a} \]

3. Table Processing:
- Convert tables to HTML format.
- Wrap the entire table with <table> and </table>.

4. Figure Handling:
- Ignore figures content in the PDF image. Do not
  attempt to describe or convert images.

5. Output Format:
- Ensure the output Markdown document has a clear
  structure with appropriate line breaks between
  elements.
- For complex layouts, try to maintain the original
  document's structure and format as closely as
  possible.

Please strictly follow these guidelines to ensure
accuracy and consistency in the conversion. Your task
is to accurately convert the content of the PDF image
into Markdown format without adding any extra
explanations or comments.
\end{lstlisting}

\subsection*{C.3\quad Inference Configuration}

All models share a common retry and timeout strategy:
\begin{itemize}
\item \textbf{Temperature:} 0.0 (deterministic decoding) for all API-based models.
\item \textbf{Max retries:} 3, with exponential backoff (5\,s, 10\,s, 20\,s).
\item \textbf{Timeout:} 120--300\,s depending on model (up to 600\,s for local models).
\item \textbf{Max image dimension:} 4000\,px (images exceeding this are resized proportionally).
\item \textbf{Max output tokens:} 8192 (where configurable).
\end{itemize}

\subsection*{C.4\quad Post-Processing}

Model outputs are processed through a \texttt{clean\_markdown()} function that strips
Markdown code fences (e.g., \texttt{```markdown ... ```}) from the response. No other
normalization is applied to model outputs before evaluation.

\subsection*{C.5\quad Human Baseline Protocol}

The human baseline is collected under conditions identical to model evaluation:
\begin{itemize}
\item Human participants receive the same handwritten document images as models.
\item \textbf{Single-pass recognition:} no revision, no second attempt.
\item \textbf{No auxiliary resources:} participants may not consult MLLM predictions,
  writer-provided transcripts, community discussions, or domain experts.
\item \textbf{Same target formats:} Markdown for text, \LaTeX{} for formulas, HTML for tables.
\item \textbf{Same evaluation metrics:} Edit Distance, CDM, and TEDS are applied identically.
\end{itemize}

This protocol yields a calibrated reference under model-comparable conditions rather than an
absolute upper bound on human performance.

\end{document}